\documentclass[final]{cvpr}

\usepackage{times}
\usepackage{epsfig}
\usepackage{graphicx}
\usepackage{amsmath}
\usepackage{amssymb}
\usepackage{cite}
\usepackage{url}            % simple URL typesetting
\usepackage{booktabs}       % professional-quality tables
\usepackage{amsfonts}       % blackboard math symbols
\usepackage{amssymb}
\usepackage{nicefrac}       % compact symbols for 1/2, etc.
\usepackage{microtype}      % microtypography
\usepackage{xcolor}
\usepackage{multirow}
\usepackage{subcaption}
\usepackage{dblfloatfix}
\usepackage{textcomp}
\usepackage[ruled,lined,linesnumbered]{algorithm2e}
\usepackage{amsmath}

\DeclareMathOperator*{\argmin}{arg\,min}

\usepackage[pagebackref=true,breaklinks=true,colorlinks,bookmarks=false]{hyperref}

\begin{document}

%%%%%%%%% TITLE
\title{Center-based 3D Object Detection and Tracking}

\newcommand{\pk}[1]{\textcolor{brown}{[PHILIPP: #1 ]}}
\newcommand{\Xingyi}[1]{\textcolor{cyan}{[Xingyi: #1 ]}}
\newcommand{\Tianwei}[1]{\textcolor{green}{[Tianwei: #1 ]}}

\newcommand{\comment}[1]{}
\newcommand{\todo}[1]{\textcolor{blue}{[TODO: #1 ]}}
\renewcommand\vec{\mathbf}

\newcommand{\lblsec}[1]{\label{sec:#1}}
\newcommand{\lblfig}[1]{\label{fig:#1}} 
\newcommand{\lbltab}[1]{\label{tbl:#1}}
\newcommand{\lbltbl}[1]{\label{tbl:#1}}
\newcommand{\lbleq}[1]{\label{eq:#1}}
\newcommand{\refsec}[1]{Section~\ref{sec:#1}}
\newcommand{\reffig}[1]{Figure~\ref{fig:#1}} 
\newcommand{\reftab}[1]{Table~\ref{tbl:#1}}
\newcommand{\reftbl}[1]{Table~\ref{tbl:#1}}
\newcommand{\refeq}[1]{Equation~\eqref{eq:#1}}
\newcommand{\refthm}[1]{Theorem~\ref{#1}}
\newcommand{\refprg}[1]{Program~\ref{#1}}
\newcommand{\refalg}[1]{Algorithm~\ref{#1}}
\newcommand{\refclm}[1]{Claim~\ref{#1}}
\newcommand{\reflem}[1]{Lemma~\ref{#1}}
\newcommand{\refpty}[1]{Property~\ref{#1}}
\newcommand{\refop}[1]{OP-\ref{op:#1}}
\newcommand{\pb}{center-based }

\author{Tianwei Yin\\
UT Austin \\ 
{\tt\small yintianwei@utexas.edu}
\and
Xingyi Zhou\\
UT Austin \\ 
{\tt\small zhouxy@cs.utexas.edu}
\and 
Philipp Kr\"ahenb\"uhl \\
UT Austin \\ 
{\tt\small philkr@cs.utexas.edu}

}

\maketitle

%%%%%%%%% ABSTRACT
\begin{abstract}
Three-dimensional objects are commonly represented as 3D boxes in a point-cloud.
This representation mimics the well-studied image-based 2D bounding-box detection but comes with additional challenges.
Objects in a 3D world do not follow any particular orientation, and box-based detectors have difficulties enumerating all orientations or fitting an axis-aligned bounding box to rotated objects.
In this paper, we instead propose to represent, detect, and track 3D objects as points.
Our framework, CenterPoint, first detects centers of objects using a keypoint detector and regresses to other attributes, including 3D size, 3D orientation, and velocity.
In a second stage, it refines these estimates using additional point features on the object.
In CenterPoint, 3D object tracking simplifies to greedy closest-point matching.
The resulting detection and tracking algorithm is simple, efficient, and effective.
CenterPoint achieved state-of-the-art performance on the nuScenes benchmark for both 3D detection and tracking, with 65.5 NDS and 63.8 AMOTA for a single model.
On the Waymo Open Dataset, CenterPoint outperforms all previous single model method by a large margin and ranks first among all Lidar-only submissions.
The code and pretrained models are available at \textcolor{magenta}{\url{https://github.com/tianweiy/CenterPoint}}.
\end{abstract}

%%%%%%%%% BODY TEXT
\section{Introduction}

Strong 3D perception is a core ingredient in many state-of-the-art driving systems~\cite{bansal2018chauffeurnet,wang2019monocular}.
Compared to the well-studied 2D detection problem, 3D detection on point-clouds offers a series of interesting challenges:
First, point-clouds are sparse, and most regions of 3D space are without measurements~\cite{hu2019exploiting}.
Second, the resulting output is a three-dimensional box that is often not well aligned with any global coordinate frame.
Third, 3D objects come in a wide range of sizes, shapes, and aspect ratios, e.g., in the traffic domain, bicycles are near planer, buses and limousines elongated, and pedestrians tall.
These marked differences between 2D and 3D detection made a transfer of ideas between the two domain harder~\cite{shi2019pointrcnn,yang20203dssd,simony2018complex}.
The crux of it is that an axis-aligned 2D box~\cite{girshick2014rich,girshick2015fast} is a poor proxy of a free-form 3D object.
One solution might be to classify a different template (anchor) for each object orientation~\cite{yang2019scrdet, yang2019r3det},
but this unnecessarily increases the computational burden and may introduce a large number of potential false-positive detections.
We argue that the main underlying challenge in linking up the 2D and 3D domains lies in this representation of objects. 

\begin{figure}[t]
   \centering
   \includegraphics[angle=0,width=\linewidth]{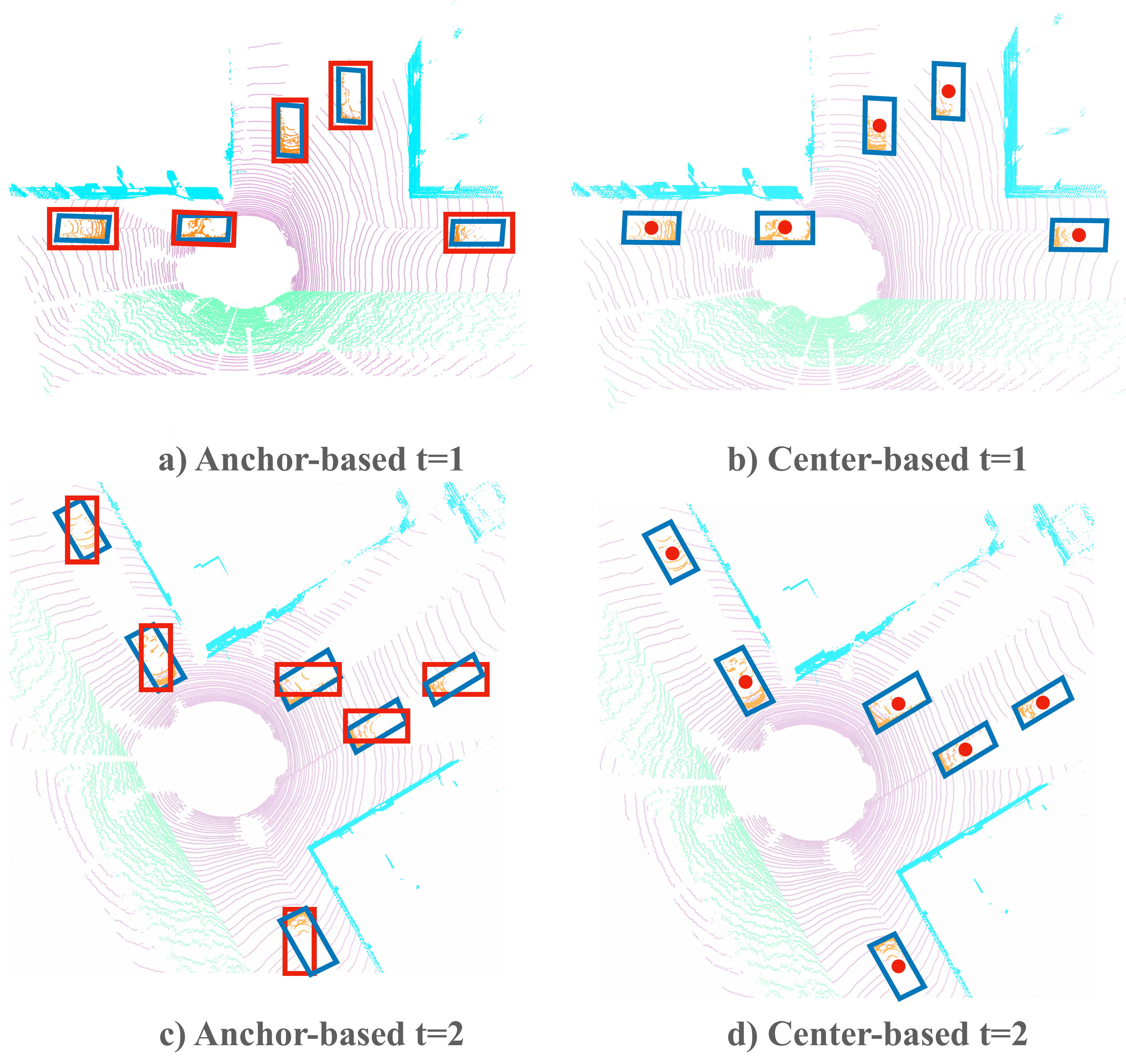}
   \caption{
    We present a center-based framework to represent, detect and track objects. Previous anchor-based methods use axis-aligned anchors with respect to ego-vehicle coordinate. When the vehicle is driving in straight roads, both anchor-based and our center-based method are able to detect objects accurately (top). However, during a safety-critical left turn (bottom), anchor-based methods have difficulty fitting axis-aligned bounding boxes to rotated objects. Our center-based model accurately detect objects through rotationally invariant points. Best viewed in color.  
   }
\lblfig{teaser}
\vspace{-3mm}
\end{figure}

In this paper, we show how representing objects as points~(\reffig{teaser}) greatly simplifies 3D recognition.
Our two-stage 3D detector, CenterPoint, finds centers of objects and their properties using a keypoint detector~\cite{zhou2019objects}, a second-stage refines all estimates.
Specifically, CenterPoint uses a standard Lidar-based backbone network, i.e., VoxelNet~\cite{voxelnet, yan2018second} or PointPillars~\cite{pillar}, to build a representation of the input point-cloud.
It then flattens this representation into an overhead map-view and uses a standard image-based keypoint detector to find object centers~\cite{zhou2019objects}.
For each detected center, it regresses to all other object properties such as 3D size, orientation, and velocity from a point-feature at the center location.
Furthermore, we use a light-weighted second stage to refine the object locations.
This second stage extracts point-features at the 3D centers of each face of the estimated objects 3D bounding box.
It recovers the lost local geometric information due to striding and a limited receptive field, and brings a decent performance boost with minor cost.

The \pb representation has several key advantages: First, unlike bounding boxes, points have no intrinsic orientation.
This dramatically reduces the object detector's search space while allowing the backbone to learn the rotational invariance of objects and rotational equivariance of their relative rotation.
Second, a \pb representation simplifies downstream tasks such as tracking.
If objects are points, tracklets are paths in space and time.
CenterPoint predicts the relative offset (velocity) of objects between consecutive frames, which are then linked up greedily.
Thirdly, point-based feature extraction enables us to design an effective two-stage refinement module that is much faster than previous approaches~\cite{pvrcnn, PartA,shi2019pointrcnn}.

We test our models on two popular large datasets: Waymo Open Dataset~\cite{sun2019scalability}, and nuScenes Dataset~\cite{caesar2019nuscenes}. 
We show that a simple switch from the box representation to center-based representation yields a $3$-$4$ mAP increase in 3D detection under different backbones ~\cite{pillar, voxelnet, yan2018second, zhu2019classbalanced}. 
Two-stage refinement further brings an additional $2$ mAP boost with small~($<10\%$) computation overhead.
Our best single model achieves $71.8$ and $66.4$ level 2 mAPH for vehicle and pedestrian detection on Waymo, $58.0$ mAP and $65.5$ NDS on nuScenes, outperforming all published methods on both datasets.   
Notably, in NeurIPS 2020 nuScenes 3D Detection challenge, CenterPoint is adopted in 3 of the top 4 winning entries. 
For 3D tracking, our model performs at $63.8$ AMOTA outperforming the prior state-of-the-art by $8.8$ AMOTA on nuScenes. 
On Waymo 3D tracking benchmark, our model achieves $59.4$ and $56.6$ level 2 MOTA for vehicle and pedestrian tracking, respectively, surpassing previous methods by up to $50\%$. 
Our end-to-end 3D detection and tracking system runs near real-time, with $11$ FPS on Waymo and $16$ FPS on nuScenes. 

\section{Related work}
\label{related_work}
\noindent
\textbf{2D object detection} predicts axis-algined bounding box from image inputs.
The RCNN family~\cite{girshick2014rich,girshick2015fast,ren2015faster,he2017mask} finds a category-agnostic bounding box candidates, then classify and refine it.
YOLO~\cite{redmon2017yolo9000}, SSD~\cite{liu2016ssd}, and RetinaNet~\cite{focal2017} directly find a category-specific box candidate, sidestepping later classification and refinement.
Center-based detectors, e.g. CenterNet~\cite{zhou2019objects} or CenterTrack~\cite{zhou2020tracking}, directly detect the implicit object center point without the need for candidate boxes.
Many 3D object detectors~\cite{shi2019pointrcnn,yang20203dssd,simony2018complex, sassd} evolved from these 2D object detector.
We argue center-based representation~\cite{zhou2019objects,zhou2020tracking} is a better fit in 3D application comparing to axis-aligned boxes.

\noindent
\textbf{3D object detection}  
aims to predict three dimensional rotated bounding boxes~\cite{yan2018second, pillar, std, qi2018frustum, yang20203dssd, liang2019multi, kitti}.
They differ from 2D detectors on the input encoder.
Vote3Deep~\cite{vote_3deep} leverages feature-centric voting~\cite{wang2015voting} to efficiently process the sparse 3D point-cloud on equally spaced 3D voxels.
VoxelNet~\cite{voxelnet} uses a PointNet~\cite{qi2017pointnet} inside each voxel to generate a unified feature representation from which a head with 3D sparse convolutions~\cite{sparse_conv} and 2D convolutions produces detections.
SECOND~\cite{yan2018second} simplifies the VoxelNet and speeds up sparse 3D convolutions. 
PIXOR~\cite{pixor} project all points onto a 2D feature map with 3D occupancy and point intensity information to remove the expensive 3D convolutions.
PointPillars~\cite{pillar} replaces all voxel computation with a pillar representation, a single tall elongated voxel per map location, improving backbone efficiency.
MVF~\cite{mvf} and Pillar-od~\cite{wang2020pillar} combine multiple view features to learn a more effective pillar representation. 
Our contribution focuses on the output representation, and is compatible with any 3D encoder and can improve them all.

VoteNet~\cite{votenet} detects objects through vote clustering using point feature sampling and grouping. 
In contrast, we directly regress to 3D bounding boxes through features at the center point without voting.  
Wong et al.~\cite{wong2019identifying} and Chen et al.~\cite{chen2019hotspot} used similar multiple points representation in the object center region (i.e., point-anchors) and regress to other attributes.
We use a single positive cell for each object and use a keypoint estimation loss.

\noindent
\textbf{Two-stage 3D object detection.} 
Recent works considered directly applying RCNN style 2D detectors to the 3D domains~\cite{pvrcnn, PartA, shi2019pointrcnn, std, fast_pointrcnn}. 
Most of them apply RoIPool~\cite{ren2015faster} or RoIAlign~\cite{he2017mask} to aggregate RoI-specific features in 3D space, using PointNet-based point~\cite{shi2019pointrcnn} or voxel~\cite{pvrcnn} feature extractor. 
These approaches extract region features from 3D Lidar measurements~(points and voxels), 
resulting in a prohibitive run-time due to massive points.
Instead, we extract sparse features of 5 surface center points from the intermediate feature map.
This makes our second stage very efficient and keeps effective.

\noindent 
\textbf{3D object tracking.} Many 2D tracking algorithms~\cite{mass, sort, deep_sort, tractor} readily track 3D objects out of the box.
However, dedicated 3D trackers based on 3D Kalman filters~\cite{weng2019baseline,chiu2020probabilistic} still have an edge as they better exploit the three-dimensional motion in a scene.
Here, we adopt a much simpler approach following CenterTrack~\cite{zhou2020tracking}.
We use a velocity estimate together with the point-based detection to track centers of objects through multiple frames.
This tracker is much faster and more accurate than dedicated 3D trackers~\cite{weng2019baseline,chiu2020probabilistic}.

\section{Preliminaries}

\noindent
\textbf{2D CenterNet}~\cite{zhou2019objects} rephrases object detection as keypoint estimation.
It takes an input image and predicts a $w \times h$ heatmap $\hat{Y} \in [0,1]^{w \times h \times K}$ for each of $K$ classes.
Each local maximum (i.e., pixels whose value is greater than its 8 neighbors) in the output heatmap corresponds to the center of a detected object.
To retrieve a 2D box, CenterNet regresses to a size map $\hat{S} \in \mathbb{R}^{w \times h \times 2}$ shared between all categories.
For each detection object, the size-map stores its width and height at the center location.
The CenterNet architecture uses a standard fully convolutional image backbone and adds a dense prediction head on top.
During training, CenterNet learns to predict heatmaps with rendered Gaussian kernels at each annotated object center $\vec{q}_i$ for each class $c_i \in \{1 \ldots K\}$, and regress to object size $S$ at the center of the annotated bounding box.
To make up for quantization errors introduced by the striding of the backbone architecture, CenterNet also regresses to a local offset $\hat{O}$.

At test time, the detector produces $K$ heatmaps and dense class-agnostic regression maps.
Each local maxima (peak) in the heatmaps corresponds to an object, with confidence proportional to the heatmap value at the peak.
For each detected object, the detector retrieves all regression values from the regression maps at the corresponding peak location.
Depending on the application domain, Non-Maxima Suppression (NMS) may be warranted.\\

\noindent 
\textbf{3D Detection}
\lblsec{3dframework}
Let $\mathcal{P} = \{(x, y, z, r)_i\}$ be an orderless point-cloud of 3D location $(x, y, z)$ and reflectance $r$ measurements.
3D object detection aims to predict a set of 3D object bounding boxes $\mathcal{B} = \{b_k\}$ in the bird eye view from this point-cloud.
Each bounding box $b = (u, v, d, w, l, h, \alpha)$ consists of a center location $(u, v, d)$, relative to the objects ground plane, and 3D size $(w, l, h)$, and rotation expressed by yaw $\alpha$.
Without loss of generality, we use an egocentric coordinate system with sensor at $(0, 0, 0)$ and yaw$=0$.

Modern 3D object detectors~\cite{voxelnet,yan2018second,pillar,sassd} uses a 3D encoder that quantizes the point-cloud into regular bins.
A point-based network~\cite{qi2017pointnet} then extracts features for all points inside a bin.
The 3D encoder then pools these features into its primary feature representation.
Most of the computation happens in the backbone network, which operates solely on these quantized and pooled feature representations.
The output of a backbone network is a map-view feature-map $\vec{M} \in \mathbb{R}^{W \times L \times F}$ of width $W$ and length $L$ with $F$ channels in a map-view reference frame.
Both width and height directly relate to the resolution of individual voxel bins and the backbone network's stride.
Common backbones include VoxelNet~\cite{voxelnet,yan2018second} and PointPillars~\cite{pillar}. 

With a map-view feature map $\vec{M}$, a detection head, most commonly a one-~\cite{focal2017} or two-stage~\cite{ren2015faster} bounding-box detector, then produces object detections from some predefined bounding boxes anchored on this overhead feature-map.
As 3D bounding boxes come with various sizes and orientation, anchor-based 3D detectors have difficulty fitting an axis-aligned 2D box to a 3D object. 
Moreover, during the training, previous anchor-based 3D detectors rely on 2D Box IoU for target assignment~\cite{yan2018second, pvrcnn}, which creates unnecessary burdens for choosing positive/negative thresholds for different classes or different dataset. 
In the next section, we show how to build a principled 3D object detection and tracking model based on point representation.  
We introduce a novel center-based detection head but rely on existing 3D backbones (VoxelNet or PointPillars). 

\begin{figure*}[t]
\centering
   \vspace{-2em}
   \includegraphics[width=0.9\linewidth]{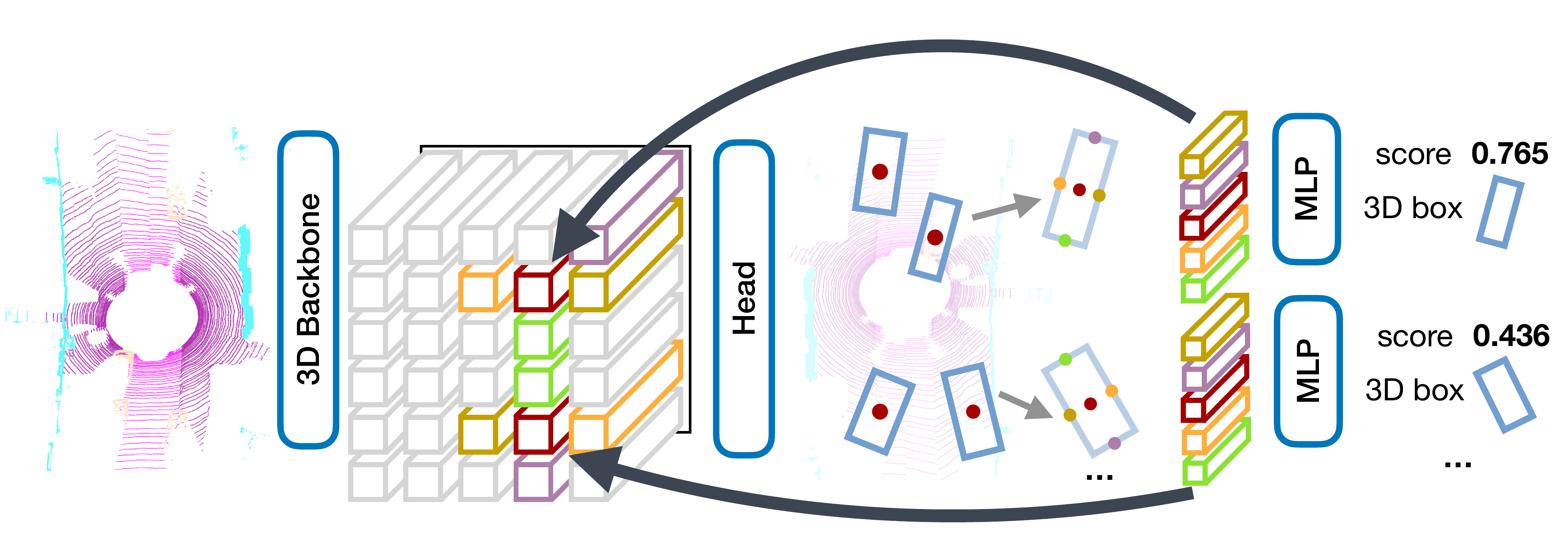}
   \begin{subfigure}[b]{.20\linewidth}\caption{Point. cloud}\lblfig{framework_pc}\end{subfigure}%
   \begin{subfigure}[b]{.24\linewidth}\caption{Map-view features}\lblfig{framework_mv}\end{subfigure}%
   \begin{subfigure}[b]{.28\linewidth}\caption{First stage:\\Centers and 3D boxes}\lblfig{framework_fs}\end{subfigure}%
   \begin{subfigure}[b]{.24\linewidth}\caption{Second stage:\\Score and 3D boxes}\lblfig{framework_ss}\end{subfigure}%
   \caption{Overview of our CenterPoint framework. We rely on a standard 3D backbone that extracts map-view feature representation from Lidar point-clouds.~Then, a 2D CNN architecture detection head finds object centers and regress to full 3D bounding boxes using center features. This box prediction is used to extract point features at the 3D centers of each face of the estimated 3D bounding box, which are passed into MLP to predict an IoU-guided confidence score and box regression refinement. Best viewed in color.}
\lblfig{framework}
\end{figure*}

\section{CenterPoint}

\reffig{framework} shows the overall framework of the CenterPoint model. 
Let $\vec{M} \in \mathbb{R}^{W\times H\times F}$ be the output of the 3D backbone.
The first stage of CenterPoint predicts a class-specific heatmap, object size, a sub-voxel location refinement, rotation, and velocity.
All outputs are dense predictions.

\noindent 
\textbf{Center heatmap head.}
The center-head's goal is to produce a heatmap peak at the center location of any detected object.
This head produces a $K$-channel heatmap $\hat Y$, one channel for each of $K$ classes.
During training, it targets a 2D Gaussian produced by the projection of 3D centers of annotated bounding boxes into the map-view. 
We use a focal loss~\cite{cornernet, zhou2019objects}.
Objects in a top-down map view are sparser than in an image.
In map-view, distances are absolute, while an image-view distorts them by perspective.
Consider a road scene, in map-view the area occupied by vehicles small, but in image-view, a few large objects may occupy most of the screen.
Furthermore, the compression of the depth-dimension in perspective projection naturally places object centers much closer to each other in image-view.
Following the standard supervision of CenterNet~\cite{zhou2019objects} results in a very sparse supervisory signal, where most locations are considered background.
To counteract this, we increase the positive supervision for the target heatmap $Y$ by enlarging the Gaussian peak rendered at each ground truth object center.
Specifically, we set the Gaussian radius to $\sigma = \max(f({wl}), \tau)$, where $\tau=2$ is the smallest allowable Gaussian radius, and $f$ is a radius function defined in CornerNet~\cite{cornernet}.
In this way, CenterPoint maintains the simplicity of the center-based target assignment; the model gets denser supervision from nearby pixels.

\noindent 
\textbf{Regression heads.}
We store several object properties at center-features of objects: a sub-voxel location refinement $o \in \mathbb{R}^2$, height-above-ground $h_g \in \mathbb{R}$, the 3D size $s \in \mathbb{R}^3$, and a yaw rotation angle $(\sin(\alpha),\cos(\alpha)) \in \mathbb{R}^2$.
The sub-voxel location refinement $o$ reduces the quantization error from voxelization and striding of the backbone network.
The height-above-ground $h_g$ helps localize the object in 3D and adds the missing elevation information removed by the map-view projection.
The orientation prediction uses the sine and cosine of the yaw angle as a continuous regression target.
Combined with box size, these regression heads provide the full state information of the 3D bounding box.
Each output uses its own head.
At training time, only ground truth centers are supervised using an L1 regression loss.
We regress to logarithmic size to better handle boxes of various shapes.
At inference time, we extract all properties by indexing into dense regression head outputs at each object's peak location.

\noindent 
\textbf{Velocity head and tracking.}
\lblsec{tracking}
To track objects through time, we learn to predict a two-dimensional velocity estimation $\vec{v} \in \mathbb{R}^2$ for each detected object as an additional regression output.
The velocity estimate is special, as it requires two input map-views the current and previous time-step.
It predicts the difference in object position between the current and the past frame.
Like other regression targets, the velocity estimation is also supervised using L1 loss at the ground truth object's location at the current time-step.

At inference time, we use this offset to associate current detections to past ones in a greedy fashion.
Specifically, we project the object centers in the current frame back to the previous frame by applying the negative velocity estimate and then matching them to the tracked objects by closest distance matching.
Following SORT~\cite{sort}, we keep unmatched tracks up to $T=3$ frames before deleting them.
We update each unmatched track with its last known velocity estimation. 
See supplement for the detailed tracking algorithm diagram.

CenterPoint combines all heatmap and regression losses in one common objective and jointly optimizes them.
It
simplifies and improves previous anchor-based 3D detectors~(see experiments).
However, all properties of the object 
are currently inferred from the object's center-feature, 
which may not contain sufficient information for accurate object localization.
For example, in autonomous driving, the sensor often only sees the side of the object, but not its center.
Next, we improve CenterPoint by using a second refinement stage with a light-weight point-feature extractor.

\subsection{Two-Stage CenterPoint}
We use CenterPoint unchanged as a first stage.
The second stage extracts additional point-features from the output of the backbone.
We extract one point-feature from the 3D center of each face of the predicted bounding box.
Note that the bounding box center, top and bottom face centers all project to the same point in map-view.
We thus only consider the four outward-facing box-faces together with the predicted object center.
For each point, we extract a feature using bilinear interpolation from the backbone map-view output $\vec M$.
Next, we concatenate the extracted point-features and pass them through an MLP.
The second stage predicts a class-agnostic confidence score and box refinement on top of one-stage CenterPoint's prediction results. 

For class-agnostic confidence score prediction, we follow ~\cite{pvrcnn, PartA, li2019gs3d, jiang2018acquisition} and use a score target $I$ guided by the box's 3D IoU with the corresponding ground truth bounding box:
\begin{equation}
   I = \min(1, \max(0, 2\times IoU_{t} - 0.5))
\end{equation}
\noindent 
where $IoU_{t}$ is the IoU between the $t$-th proposal box and the ground-truth. 
The training is supervised with a binary cross entropy loss:
\begin{equation}
    L_{score} =  -I_t \log(\hat{I_t}) - (1-I_t)\log(1-\hat{I_t}) 
\end{equation}
\noindent 
where $\hat{I_t}$ is the predicted confidence score. 
During the inference, we directly use the class prediction from one-stage CenterPoint and computes the final confidence score as the geometric average of the two scores $\hat{Q}_{t} = \sqrt{\hat{Y}_{t} * \hat{I}_t}$
where $\hat{Q}_t$ is the final prediction confidence of object $t$ and $\hat{Y}_t =  \max_{0 \leq k \leq K} \hat{Y}_{p,k}$ and $\hat{I}_t$ are the first stage and second stage confidence of object $t$, respectively. 

For box regression, the model predicts a refinement on top of first stage proposals, and we train the model with L1 loss.
Our two-stage CenterPoint simplifies and accelerates previous two-stage 3D detectors that use expensive PointNet-based feature extractor and RoIAlign operations~\cite{shi2019pointrcnn, pvrcnn}.

\subsection{Architecture}
All first-stage outputs share a first $3 \times 3$ convolutional layer, Batch Normalization~\cite{ioffe2015batch}, and ReLU.
Each output then uses its own branch of two $3 \times 3$ convolutions separated by a batch norm and ReLU.
Our second-stage uses a shared two-layer MLP, with a batch norm, ReLU, and Dropout~\cite{hinton2012improving} with a drop rate of $0.3$, followed by two branches of three fully-connected layers, one for confidence score and one for box regression prediction. 

\section{Experiments}

We evaluate CenterPoint on Waymo Open Dataset and nuScenes dataset.
We implement CenterPoint using two 3D encoders: VoxelNet~\cite{voxelnet,yan2018second,zhu2019classbalanced} and PointPillars~\cite{pillar}, termed CenterPoint-Voxel and CenterPoint-Pillar respectively.\\

\noindent 
\textbf{Waymo Open Dataset.}
Waymo Open Dataset~\cite{sun2019scalability} contains 798 training sequences and 202 validation sequences for vehicle and pedestrian. 
The point-clouds are captured with a 64 lanes Lidar, which produces about 180k Lidar points every 0.1s. 
The official 3D detection evaluation metrics include the standard 3D bounding box mean average precision (mAP) and mAP weighted by heading accuracy (mAPH).
The mAP and mAPH are based on an IoU threshold of 0.7 for vehicles and 0.5 for pedestrians. 
For 3D tracking, 
the official metrics are
Multiple Object Tracking Accuracy~(MOTA) and Multiple Object Tracking Precision~(MOTP)~\cite{bernardin2006multiple}.
The official evaluation toolkit also provides a performance breakdown for two difficulty levels: 
LEVEL\_1 for boxes with more than five Lidar points, and LEVEL\_2 for boxes with at least one Lidar point. 

Our Waymo model uses a detection range of $[-75.2\text{m}, 75.2\text{m}]$ for the $X$ and $Y$ axis, and $[-2\text{m}, 4\text{m}]$ for the $Z$ axis. 
CenterPoint-Voxel uses a $(0.1\text{m}, 0.1\text{m}, 0.15\text{m})$ voxel size following PV-RCNN~\cite{pvrcnn} while CenterPoint-Pillar uses a grid size of $(0.32\text{m}, 0.32\text{m})$. 

\paragraph{nuScenes Dataset.}
nuScenes~\cite{caesar2019nuscenes} contains 1000 driving sequences, with 700, 150, 150 sequences for training, validation, and testing, respectively.
Each sequence is approximately 20-second long, with a Lidar frequency of 20 FPS.
The dataset provides calibrated vehicle pose information for each Lidar frame but only provides box annotations every ten frames (0.5s).
nuScenes uses a 32 lanes Lidar, which produces approximately 30k points per frame. 
In total, there are 28k, 6k, 6k, annotated frames for training, validation, and testing, respectively.
The annotations include 10 classes with a long-tail distribution. 
The official evaluation metrics are an average among the classes.
For 3D detection, the main metrics are mean Average Precision (mAP)~\cite{everingham2010pascal} and nuScenes detection score (NDS).
The mAP uses a bird-eye-view center distance $<{0.5\text{m}, 1\text{m}, 2\text{m}, 4\text{m}}$ instead of standard box-overlap.
NDS is a weighted average of mAP and other attributes metrics, including translation, scale, orientation, velocity, and other box attributes~\cite{caesar2019nuscenes}.
After our test set submission, the nuScenes team adds a new neural planning metric~(PKL)~\cite{philion2020learning}.  
The PKL metric measures the influence of 3D object detection for down-streamed autonomous driving tasks based on the KL divergence of a planner's route~(using 3D detection) and the ground truth trajectory.
Thus, we also report the PKL metric for all methods that evaluate on the test set.

For 3D tracking, nuScenes uses AMOTA~\cite{weng2019baseline}, which penalizes ID switches, false positive, and false negatives and is averaged among various recall thresholds. 

For experiments on nuScenes, we set the detection range to $[-51.2\text{m}, 51.2\text{m}]$ for the $X$ and $Y$ axis, and $[-5\text{m}, 3\text{m}]$ for $Z$ axis.
CenterPoint-Voxel use a $(0.1\text{m}, 0.1\text{m}, 0.2\text{m})$ voxel size and CenterPoint-Pillars uses a $(0.2\text{m}, 0.2\text{m})$ grid. \\ 

\noindent 
\textbf{Training and Inference.}
We use the same network designs and training schedules as prior works~\cite{pvrcnn, zhu2019classbalanced}.
See supplement for detailed hyper-parameters. % also include network details  
During the training of two-stage CenterPoint, we randomly sample $128$ boxes with $1$:$1$ 
positive negative ratio~\cite{ren2015faster} from the first stage predictions. 
A proposal is positive if it overlaps with a ground truth annotation with at least 0.55 IoU~\cite{pvrcnn}.
During inference, we run the second stage on the top 500 predictions after Non-Maxima Suppression~(NMS).
The inference times are measured on an Intel Core i7 CPU and a Titan RTX GPU.

{
\begin{table}[t]
\small
\begin{center}
\begin{tabular}{@{}l@{\ \ }l@{\ \ }c@{\ \ }c@{\ \ }c@{\ \ }c@{}}
  \toprule 
   \multirow{2}{4em}{Difficulty} & \multirow{2}{4em}{Method} &  \multicolumn{2}{c}{Vehicle} & \multicolumn{2}{c}{Pedestrian}  \\
   & & mAP & mAPH &  mAP & mAPH \\
    \cmidrule(r){1-2}
    \cmidrule(r){3-4}
    \cmidrule(r){5-6}
    \multirow{5}{3em}{Level 1}
  & StarNet \cite{ngiam2019starnet} & 61.5 & 61.0 & 67.8 & 59.9  \\   
  & PointPillars \cite{pillar} & 63.3 & 62.8 & 62.1 & 50.2   \\  
  & PPBA \cite{ngiam2019starnet} & 67.5 & 67.0 & 69.7 & 61.7   \\   
  & RCD \cite{bewley2020range} & 72.0 & 71.6 & \_ & \_  \\ 
  & Ours & \textbf{80.2} & \textbf{79.7} & \textbf{78.3} & \textbf{72.1} \\
    \cmidrule(r){1-2}
    \cmidrule(r){3-4}
    \cmidrule(r){5-6}
    \multirow{5}{3em}{Level 2}
  & StarNet \cite{ngiam2019starnet} & 54.9 & 54.5 & 61.1 & 54.0   \\   
  & PointPillars \cite{pillar} & 55.6 & 55.1 & 55.9 & 45.1   \\  
  & PPBA \cite{ngiam2019starnet} & 59.6 & 59.1 & 63.0 & 55.8  \\   
  & RCD \cite{bewley2020range} & 65.1 & 64.7 & \_ & \_ \\  
  & Ours & \textbf{72.2} & \textbf{71.8} & \textbf{72.2} & \textbf{66.4}  \\ 

\bottomrule
 \end{tabular}
\end{center}
\vspace{-5mm}
\small
\caption{State-of-the-art comparisons for 3D detection on Waymo test set. We show the mAP and mAPH for both level 1 and level 2 benchmarks.}
\lbltab{waymo_detection}
\end{table}
}

{
\begin{table}[t]
\small
\begin{center}
\begin{tabular}{lccc}
  \toprule 
  Method  & mAP$\uparrow$ & NDS$\uparrow$ & PKL$\downarrow$ \\ 
  \midrule
  WYSIWYG \cite{hu2019exploiting}  & 35.0 & 41.9 & 1.14 \\ 
  PointPillars \cite{pillar} & 40.1 & 55.0 & 1.00 \\ 
  CVCNet \cite{chen2020view}  & 55.3 & 64.4 & 0.92 \\ 
  PointPainting \cite{vora2019pointpainting}  & 46.4 & 58.1 & 0.89 \\ 
  PMPNet \cite{yin2020Lidarbased} & 45.4 & 53.1 & 0.81 \\ 
  SSN \cite{zhu2020ssn} & 46.3 & 56.9 & 0.77 \\ 
  CBGS \cite{zhu2019classbalanced} & 52.8 &  63.3 & 0.77 \\   
 \midrule 
  Ours & \textbf{58.0} & \textbf{65.5} & \textbf{0.69} \\  
  \bottomrule
 \end{tabular}
\end{center}
\vspace{-5mm}
\small
\caption{State-of-the-art comparisons for 3D detection on nuScenes test set. We show the nuScenes detection score~(NDS), and mean Average Precision~(mAP). }
\lbltab{nuscenes_detection}
\vspace{-3mm}
\end{table}
}

\subsection{Main Results}

\paragraph{3D Detection} 
We first present our 3D detection results on the test sets of Waymo and nuScenes.
Both results use a single CenterPoint-Voxel model. 
\reftab{waymo_detection} and \reftab{nuscenes_detection} summarize our results.
On Waymo test set, our model achieves $71.8$ level 2 mAPH for vehicle detection and $66.4$ level 2 mAPH for pedestrian detection, surpassing previous methods by $7.1\%$ mAPH for vehicles and $10.6\%$ mAPH for pedestrians.
On nuScenes~(\reftab{nuscenes_detection}), our model outperforms the last-year challenge winner CBGS~\cite{zhu2019classbalanced} with multi-scale inputs and multi-model ensemble by $5.2\%$ mAP and $2.2\%$ NDS.
Our model is also much faster, as shown later.
A breakdown along classes is contained in the supplementary material.
Our model displays a consistent performance improvement over all categories and shows more significant improvements in small categories ($+5.6$ mAP for traffic cone) and extreme-aspect ratio categories ($+6.4$ mAP for bicycle and $+7.0$ mAP for construction vehicle).
More importantly, our model significantly outperforms all other submissions under the neural planar metric (PKL), a hidden metric evaluated by the organizers after our leaderboard submission.
This highlights the generalization ability of our framework.

\begin{table}[t]
\small
\begin{center}
\begin{tabular}{@{}l@{\ \ \ }l@{\ \ \ }c@{\ \ \ }c@{\ \ \ }c@{\ \ \ }c@{}}
  \toprule
  \multirow{2}{4em}{Difficulty} & \multirow{2}{3em}{Method} &  \multicolumn{2}{c}{MOTA$\uparrow$} & \multicolumn{2}{c}{MOTP$\downarrow$} \\ 
  & & Vehicle & Ped. & Vechile & Ped.  \\ 
   \midrule 
   \multirow{2}{3em}{Level 1} 
   & AB3D~\cite{weng2019baseline, sun2019scalability} & 42.5  & 38.9 & 18.6 & 34.0  \\
   & Ours &  \textbf{62.6} & \textbf{58.3} & \textbf{16.3} & \textbf{31.1} \\ 
  \midrule 
  \multirow{2}{3em}{Level 2} 
  & AB3D~\cite{weng2019baseline, sun2019scalability} & 40.1 & 37.7 & 18.6 & 34.0  \\ 
  & Ours & \textbf{59.4} & \textbf{56.6} & \textbf{16.4} & \textbf{31.2} \\ 
  \bottomrule
\end{tabular}
\end{center}
\vspace{-5mm}
\caption{State-of-the-art comparisons for 3D tracking on Waymo test set. We show MOTA, and MOTP. $\uparrow$ is for higher better and $\downarrow$ is for lower better.}
\lbltab{track_waymo}
\end{table}

\begin{table}[t]
\small
\begin{center}
\begin{tabular}{lcccc}
  \toprule
  Method & AMOTA$\uparrow$ & FP$\downarrow$ & FN$\downarrow$ & IDS$\downarrow$\\
   \midrule 
  AB3D \cite{weng2019baseline} & 15.1 & \textbf{15088} & 75730 & 9027 \\
  Chiu et al. \cite{chiu2020probabilistic} & 55.0  & 17533 & 33216 & 950\\
  Ours & \textbf{63.8} & 18612 & \textbf{22928} & \textbf{760} \\ 
  \bottomrule
\end{tabular}
\end{center}
\vspace{-5mm}
\caption{State-of-the-art comparisons for 3D tracking on nuScenes test set. We show AMOTA, the number of false positives (FP), false negatives (FN), id switches (IDS), and per-category AMOTA. $\uparrow$ is for higher better and $\downarrow$ is for lower better.}
\lbltab{track_nusc}
\end{table}

\paragraph{3D Tracking} 
\reftab{track_waymo} shows CenterPoint's tracking performance on the Waymo test set. 
Our velocity-based closest distance matching described in \refsec{tracking} significantly outperforms the official tracking baseline in the Waymo paper~\cite{sun2019scalability}, which uses a Kalman-filter based tracker~\cite{weng2019baseline}.
We observe a $19.4$ and $18.9$ MOTA improvement for vehicle and pedestrian tracking, respectively.
On nuScenes (\reftab{track_nusc}), our framework outperforms the last challenge winner Chiu et al. \cite{chiu2020probabilistic} by $8.8$ AMOTA. 
Notably, our tracking does not require a separate motion model and runs in a negligible time, $1ms$ on top of detection.

{
\begin{table}[t]
\small
\begin{center}
\begin{tabular}{l@{\ \ }c@{\ \ }c@{\ \ }c@{\ \ }c}
  \toprule 
  Encoder & Method & Vehicle & Pedestrain & mAPH \\
  \midrule 
    \multirow{2}{5em}{VoxelNet}
   & Anchor-based & 66.1 & 54.4 & 60.3  \\
   & Center-based &  \textbf{66.5} & \textbf{62.7} & \textbf{64.6}  \\
  \midrule
    \multirow{2}{5em}{PointPillars}
  & Anchor-based  &  64.1 & 50.8 & 57.5   \\ 
  & Center-based &  \textbf{66.5} & \textbf{57.4} & \textbf{62.0}  \\
\bottomrule
 \end{tabular}
\end{center}
\vspace{-5mm}
\caption{Comparison between anchor-based and center-based methods for 3D detection on Waymo validation. We show the per-calss and average LEVEL\_2 mAPH.}
\lbltab{waymo_first}
\end{table}
}

{
\begin{table}[t]
\small
\begin{center}
\begin{tabular}{l@{\ \ }c@{\ \ }c@{\ \ }c}
  \toprule 
  Encoder & Method & mAP & NDS \\
  \midrule 
    \multirow{2}{5em}{VoxelNet}
   & Anchor-based & 52.6 & 63.0 \\
   & Center-based & \textbf{56.4} & \textbf{64.8}    \\
  \midrule
    \multirow{2}{5em}{PointPillars}
  & Anchor-based  & 46.2  & 59.1  \\ 
  & Center-based & \textbf{50.3} & \textbf{60.2}  \\
\bottomrule
 \end{tabular}
\end{center}
\vspace{-5mm}
\caption{Comparison between anchor-based and center-based methods for 3D detection on nuScenes validation. We show mean average precision (mAP) and nuScenes detection score (NDS).}
\lbltab{nusc_det}
\vspace{-2mm}
\end{table}
}

\subsection{Ablation studies}
\label{ablation}

\noindent 
\textbf{Center-based vs~Anchor-based}
We first compare our center-based one-stage detector with its anchor-based counterparts~\cite{yan2018second, zhu2019classbalanced, pillar}. 
On Waymo, we follow the state-of-the-art PV-RCNN~\cite{pvrcnn} to set the anchor hyper-parameters: we use two anchors per-locations with $0$\textdegree and $90$\textdegree; The positive/ negative IoU thresholds are set as $0.55/ 0.4$ for vehicles and $0.5/ 0.35$ for pedestrians.
On nuScenes, we follow the anchor assignment strategy from the last challenge winner CBGS~\cite{zhu2019classbalanced}. 
All other parameters are the same as our CenterPoint model. 

As is shown in \reftab{waymo_first}, on Waymo dataset, simply
switching from anchors to our centers 
gives $4.3$ mAPH and $4.5$ mAPH improvements for VoxelNet and PointPillars encoder, respectively.
On nuScenes (\reftab{nusc_det}) CenterPoint improves anchor-based counterparts by $3.8$-$4.1$ mAP and $1.1$-$1.8$ NDS across different backbones.
To understand where the improvements are from, 
we further show the performance breakdown on different subsets based on object sizes and orientation angles on the Waymo validation set.

{
\begin{table}[t]
\small
\begin{center}
\begin{tabular}{@{}l@{\ }c@{\ }c@{\ }c@{\ }c@{\ }c@{\ }c@{\ }c@{}}
  \toprule 
   &  \multicolumn{3}{c}{Vehicle} & \multicolumn{3}{c}{Pedestrian} \\ 
   Rel. yaw & 0\textdegree-15\textdegree & 15\textdegree-30\textdegree & 30\textdegree-45\textdegree & 0\textdegree-15\textdegree & 15\textdegree-30\textdegree & 30\textdegree-45\textdegree\\ 
  \# annot. & 81.4\% & 10.5\% & 8.1\% & 71.4\% & 15.8\% & 12.8\% \\ 
\midrule 
    Anchor-based & 67.1 & \textbf{47.7} & 45.4 & 55.9 & 32.0 & 26.5  \\
    Center-based & \textbf{67.8} & 46.4 & \textbf{51.6} & \textbf{64.0} & \textbf{42.1} & \textbf{35.7}\\ 
\bottomrule
 \end{tabular}
\end{center}
\vspace{-5mm}
\caption{Comparison between anchor-based and center based methods for detecting objects of different heading angles. The ranges of the rotation angle and their corresponding portion of objects are listed in line 2 and line 3. We show the LEVEL\_2 mAPH for both methods on the Waymo validation. }
\lbltab{orientation_ablation}
\end{table}
}

{
\begin{table}[t]
\small
\begin{center}
\begin{tabular}{@{}l@{\ \ }c@{\ \ }c@{\ \ }c@{\ \ }c@{\ \ }c@{\ \ }c@{\ \ }c@{}}
  \toprule 
  \multirow{2}{4em}{Method} &  \multicolumn{3}{c}{Vehicle} & \multicolumn{3}{c}{Pedestrian} \\ 
   & small & medium & large & small & medium & large\\ 
\midrule 
    Anchor-based & 58.5 & \textbf{72.8} & 64.4 & 29.6 & 60.2 & 60.1  \\
    Center-based & \textbf{59.0} & 72.4 & \textbf{65.4} & \textbf{38.5} & \textbf{69.5} & \textbf{69.0} \\ 
\bottomrule
 \end{tabular}
\end{center}
\vspace{-5mm}
\caption{Effects of object size for the performance of anchor-based and center-based methods. We show the per-class LEVEL\_2 mAPH for objects in different size range:~{small 33\%, middle 33\%, and large 33\%}}
\lbltab{size_ablation}
\vspace{-3mm}
\end{table}
}

We first divide the ground truth instances into three bins based on their heading angles: 0\textdegree to 15\textdegree, 15\textdegree to 30\textdegree, and 30\textdegree to 45\textdegree.
This division tests the detector's performance for detecting heavily rotated boxes, which is critical for the safe deployment of autonomous driving. 
We also divide the dataset into three splits: small, medium, and large, and each split contains $\frac{1}{3}$ of the overall ground truth boxes.  

\reftab{orientation_ablation} and \reftab{size_ablation} summarize the results. 
Our center-based detectors perform much better than the anchor-based baseline when the box is rotated or deviates from the average box size, demonstrating the model's ability to capture the rotation and size invariance when detecting objects. 
These results convincingly highlight the advantage of using a point-based 
representation of 3D objects.\\

\noindent 
\textbf{One-stage vs. Two-stage}
In \reftab{waymo_second}, we show the comparison between single and two-stage CenterPoint models using 2D CNN features on Waymo validation. 
Two-stage refinement with multiple center features gives a large accuracy boost to both 3D encoders with small overheads (6ms-7ms).
We also compare with RoIAlign, which densely samples $6\times6$ points in the RoI~\cite{PartA, pvrcnn}, our center-based feature aggregation achieved comparable performance but is faster and simpler. 

The voxel quantization limits two-stage CenterPoint's improvements for pedestrian detection with PointPillars as pedestrians typically only reside in 1 pixel in the model input. 

Two-stage refinement does not bring an improvement over the single-stage CenterPoint model on nuScenes in our experiments. 
We think the reason is that the nuScenes dataset uses 32 lanes Lidar, which produces about 30k Lidar points per frame, about $\frac{1}{6}$ of the number of points in the Waymo dataset, which limits the potential improvements of two-stage refinement. 
Similar results have been observed in previous two-stage methods like PointRCNN~\cite{shi2019pointrcnn} and PV-RCNN~\cite{pvrcnn}.

\begin{table}[t]
\small
\vspace{-2mm}
\begin{center}
\begin{tabular}{@{}l@{\ }l@{\ }c@{\ \ }c@{\ }c@{\ }c@{}}
  \toprule
   Encoder & Method & Vehicle & Ped. & $T_{proposal}$ & $T_{refine}$ \\
  \midrule
    \multirow{3}{5em}{VoxelNet}
   & First Stage & 66.5 & 62.7 & 71ms & \_ \\
   & + Box Center & 68.0 & 64.9 & 71ms & 5ms  \\
   & + Surface Center & \textbf{68.3} & 65.3 & 71ms & 6ms  \\ 
   & Dense Sampling & 68.2 & \textbf{65.4}  & 71ms & 8ms \\ 
   \midrule
    \multirow{3}{5em}{PointPillars}
   & First Stage & 66.5 & 57.4 & 56ms & \_ \\
   & + Box Center & 67.3 & 57.4 & 56ms & 6ms  \\
   & + Surface Center & \textbf{67.5} & 57.9 & 56ms & 7ms  \\ 
   & Dense Sampling & 67.3 & \textbf{57.9} & 56ms & 8ms  \\ 
   \bottomrule

\end{tabular}
\end{center}
\vspace{-5mm}
\caption{Compare 3D LEVEL\_2 mAPH for VoxelNet and PointPillars encoders using single stage, two stage with 3D center features, and two stage with 3D center and surface center features on Waymo validation. 
}
\lbltab{waymo_second}
\end{table}

\begin{table}[t]
\small
\begin{center}
\begin{tabular}{@{}l@{\ \ }c@{\ \ }c@{\ \ }c@{}}
  \toprule
   Methods & Vehicle & Pedestrian  & Runtime \\
  \midrule
  BEV Feature  & 68.3 & 65.3 & 77ms  \\ 
  w/ VSA~\cite{pvrcnn} & 68.3 & 65.2 &  98ms \\ 
  w/ RBF Interpolation~\cite{sassd, qi2017pointnet++} & \textbf{68.4} & \textbf{65.7} & 89ms  \\ 
  \bottomrule
\end{tabular}
\end{center}
\vspace{-5mm}
\caption{ Ablation studies of different feature components for two stage refinement module. VSA stands for Voxel Set Abstraction, the feature aggregation methods used in PV-RCNN~\cite{pvrcnn}. RBF uses a radial basis function to interpolate 3 nearest neighbors. We compare bird-eye view and 3D voxel features using LEVEL\_2 mAPH on Waymo validation. 
}
\vspace{-3mm}
\lbltab{waymo_voxel}
\end{table}

{
\begin{table}[t]
\small
\begin{center}
\begin{tabular}{@{}l@{\ \ }l@{\ \ }c@{\ \ }c@{\ \ }c@{\ \ }c@{}}
  \toprule 
   \multirow{2}{4em}{Difficulty} & \multirow{2}{4em}{Method} &  \multicolumn{2}{c}{Vehicle} & \multicolumn{2}{c}{Pedestrian}  \\
   & & mAP & mAPH &  mAP & mAPH \\
       \cmidrule(r){1-2}
    \cmidrule(r){3-4}
    \cmidrule(r){5-6}
  \multirow{7}{3em}{Level 1}  
  & DOPS \cite{najibi2020dops} & 56.4 & \_ & \_ \\ 
  & PointPillars \cite{pillar}  & 56.6 & \_  &  59.3 & \_  \\  
  & PPBA \cite{ngiam2019starnet} & 62.4 & \_ & 66.0 & \_  \\   
  & MVF \cite{mvf}  & 62.9 & \_ & 65.3& \_  \\ 
  & Huang et al. \cite{huang2020lstm}  & 63.6 & \_ & \_  \\
  & AFDet \cite{ge2020afdet} & 63.7 & \_ & \_ \\ 
  & CVCNet \cite{chen2020view} & 65.2 & \_ & \_ \\ 
  & Pillar-OD \cite{wang2020pillar}  & 69.8 & \_ & 72.5& \_  \\ 
  & PV-RCNN \cite{pvrcnn}  & 74.4 & 73.8 & 61.4& 53.4  \\
  & CenterPoint-Pillar(ours) & 76.1 & 75.5 & 76.1 & 65.1  \\ 
  & CenterPoint-Voxel(ours) & \textbf{76.7} & \textbf{76.2} & \textbf{79.0} & \textbf{72.9}  \\
  \midrule
  \multirow{2}{3em}{Level 2}  
  & PV-RCNN \cite{pvrcnn} & 65.4 & 64.8 & 53.9 & 46.7 \\ 
  & CenterPoint-Pillar(ours) & 68.0 & 67.5 & 68.1 & 57.9 \\ 
  & CenterPoint-Voxel(ours) &  \textbf{68.8} & \textbf{68.3} & \textbf{71.0} & \textbf{65.3} \\ 
  \bottomrule
 
 \end{tabular}
\end{center}
\vspace{-5mm}
\caption{State-of-the-art comparisons for 3D detection on Waymo validation.}
\lbltab{waymo_val}
\end{table}
}

\begin{table}[t]
\small
\begin{center}
\begin{tabular}{@{}l@{\ }c@{\ }c@{\ }c@{\ }c@{\ }c@{}}
  \toprule
  Detector & Tracker & AMOTA$\uparrow$ & AMOTP$\downarrow$  & $T_{track} $ & $T_{tot}$\\
  \midrule
 CenterPoint-Voxel & Point & \textbf{63.7} & \textbf{0.606} & 1ms & 62ms\\ 
 CBGS~\cite{zhu2019classbalanced} & Point & 59.8 & 0.682 & 1ms & $>182ms$ \\
 CenterPoint-Voxel & M-KF & 60.0 & 0.765 & 73ms & 135ms   \\ 
  CBGS~\cite{zhu2019classbalanced} & M-KF & 56.1 & 0.800 & 73ms & $>$254ms \\
  \bottomrule 
\end{tabular}
\end{center}
\vspace{-5mm}
\caption{Ablation studies for 3D tracking on nuScenes validation. We show combinations of different detectors and trackers. CenterPoint-* are our detectors. Point is our proposed tracker. M-KF is short for Mahalanobis distance-based Kalman filter, as is used in the last challenge winner Chiu et al.~\cite{chiu2020probabilistic}. $T_{track}$ denotes tracking time and $T_{tot}$ denotes total time for both detection and tracking.}
\lbltab{track_ablation}
\vspace{-5mm}
\end{table}

\begin{figure*}[t]
\small 
\centering 
\includegraphics[trim=0 95 0 100, clip, width=0.45\linewidth]{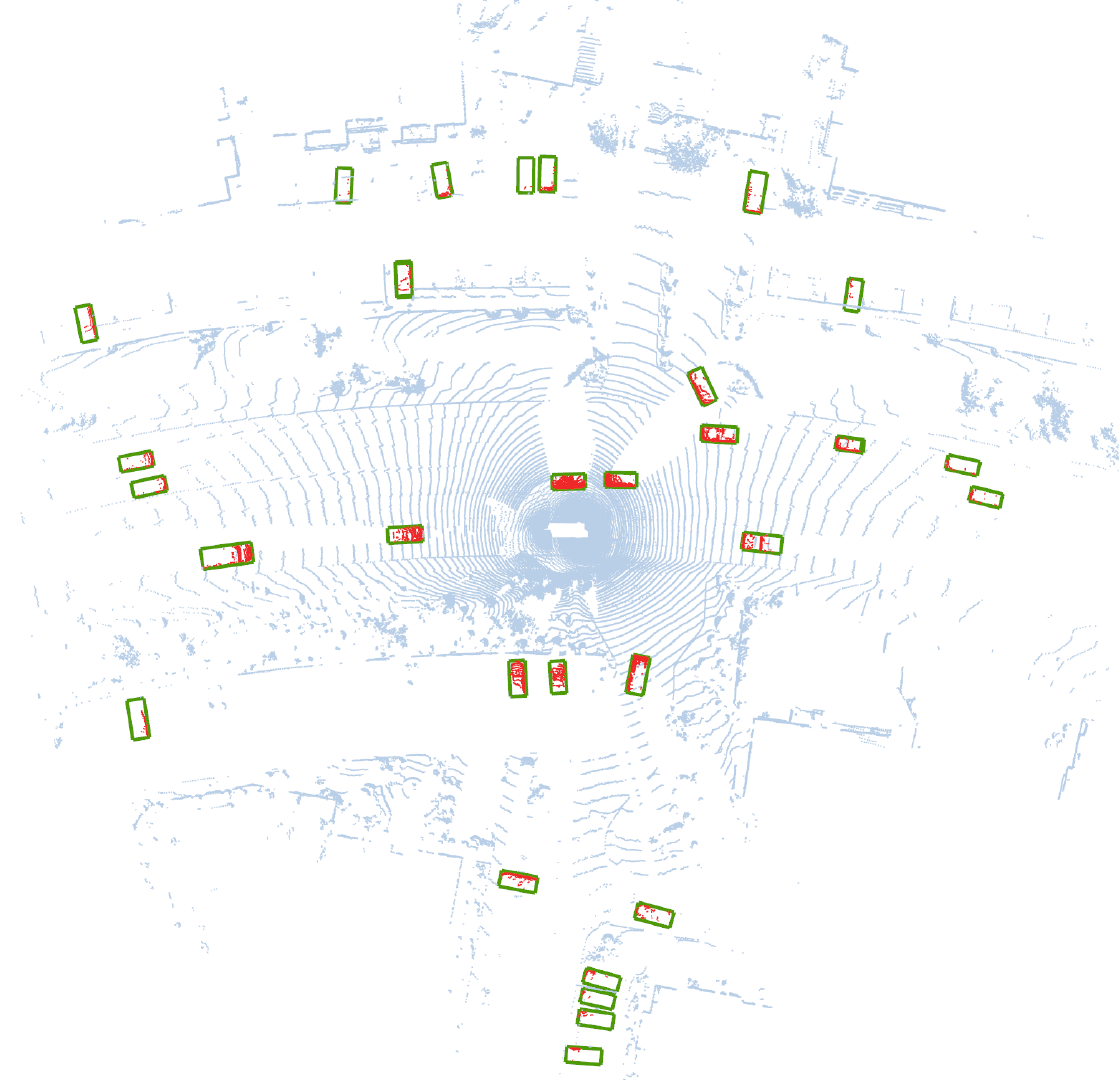} 
\includegraphics[trim=0 95 0 95, clip, width=0.45\linewidth]{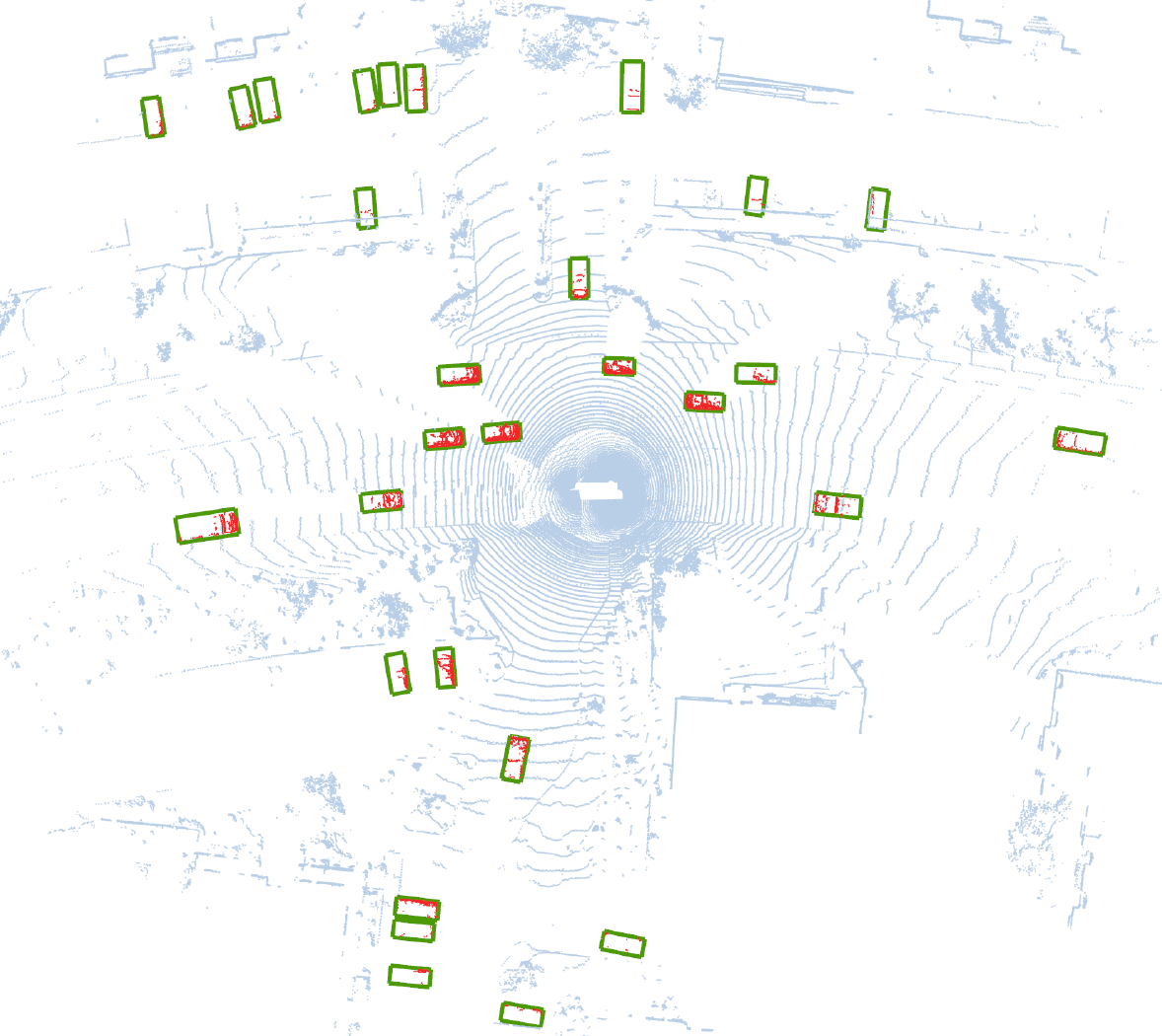} 
\caption{Example qualitative results of CenterPoint on the Waymo validation. We show the raw point-cloud in blue, our detected objects in green bounding boxes, and Lidar points inside bounding boxes in red. Best viewed on screen. }
\lblfig{visualization}
\end{figure*}

\noindent 
\textbf{Effects of different feature components}
In our two-stage CenterPoint model, we only use features from the 2D CNN feature map. 
However, previous methods propose to also utilize voxel features for second stage refinement~\cite{pvrcnn, PartA}. 
Here, we compare with two voxel feature extraction baselines:

\textbf{Voxel-Set Abstraction.}
PV-RCNN~\cite{pvrcnn} proposes the Voxel-Set Abstraction (VSA) module, which extends PointNet++~\cite{qi2017pointnet++}'s set abstraction layer to aggregate voxel features in a fixed radius ball.

\textbf{Radial basis function (RBF) Interpolation.}
PointNet++\cite{qi2017pointnet++} and SA-SSD\cite{sassd} use a radial basis function to aggregate grid point features from three nearest non-empty 3D feature volumes. 

For both baselines, we combine bird-eye view features with voxel features using their official implementations.  
\reftab{waymo_voxel} summarizes the results. 
It shows bird-eye view features are sufficient for good performance while being more efficient comparing to voxel features used in the literatures~\cite{pvrcnn,sassd,qi2017pointnet++}.

To compare with prior work that did not evaluate on Waymo test, we also report results on the Waymo validation split in \reftab{waymo_val}.
Our model outperforms all published methods by a large margin, especially for the challenging pedestrian class(+18.6 mAPH) of the level 2 dataset, where boxes contain as little as one Lidar point.   

\noindent 
\textbf{3D Tracking.} 
\reftab{track_ablation} shows the ablation experiments of 3D tracking on nuScenes validation.  
We compare with last year's challenge winner Chiu et al.~\cite{chiu2020probabilistic}, which uses mahalanobis distance-based Kalman filter to associate detection results of CBGS~\cite{zhu2019classbalanced}.  
We decompose the evaluation into the detector and tracker to make the comparison strict. 
Given the same detected objects, using our simple velocity-based closest point distance matching outperforms the Kalman filter-based Mahalanobis distance matching~\cite{chiu2020probabilistic} by $3.7$ AMOTA (line 1 vs. line 3 and line 2 vs. line4). 
There are two sources of improvements: 1) we model the object motion with a learned point velocity, rather than modeling 3D bounding box dynamic with a Kalman filter; 2) we match objects by center point-distance instead of a Mahalanobis distance of box states or 3D bounding box IoU. 
More importantly, our tracking is a simple nearest neighbor matching without any hidden-state computation.
This saves the computational overhead of a 3D Kalman filter~\cite{chiu2020probabilistic} (73ms vs. 1ms).

\section{Conclusion}

We proposed a center-based framework for simultaneous 3D object detection and tracking from the Lidar point-clouds. 
Our method uses a standard 3D point-cloud encoder with a few convolutional layers in the head to produce a bird-eye-view heatmap and other dense regression outputs.
Detection is a simple local peak extraction with refinement, and tracking is a closest-distance matching.
CenterPoint is simple, near real-time, and achieves state-of-the-art performance on the Waymo and nuScenes benchmarks.

% -------------
{\small
\bibliographystyle{ieee_fullname}
\bibliography{egbib}
}

\clearpage
\appendix
\section{Tracking algorithm}

\begin{algorithm}
    \caption{{Center-based Tracking}}
	\label{alg:association}
	\SetAlgoLined
	\SetKwInOut{Input}{Input} \SetKwInOut{Output}{Output} 
    \Input{$T^{(t - 1)} = \{(\vec{p}, \vec{v}, {c}, \vec{q}, id, a)_j^{(t-1)}\}_{j=1}^{M}$:
    Tracked objects in the previous frame, with center $\vec p$, ground plane velocity $\vec v$, category label $c$, other bounding box attributes $\vec{q}$, tracking id $id$, and inactive age $a$ (active tracks will have $a=0$). \\
    $\hat{D}^{(t)} = \{(\hat{\vec p}, \hat{\vec v}, \hat{c}, \hat{\vec q})_i^{(t)}\}_{i=1}^{N}$: Detections in the current frame in descending confidence.
    \\ }
	\Output{
	        $T^{(t)} = \{(\vec{p}, \vec{v}, {c}, \vec{q}, id, a)_{j=1}^{K} \}$: Tracked Objects.
	}
    \textbf{Hyper parameters:} Matching distance threshold $\tau$; Max inactive age $A$.\\
    \textbf{Initialization:} Tracks $T^{(t)}$, and matches $\mathcal{S}$ are initialized as empty sets. \label{alg:st} \\
    $T^{(t)} \leftarrow \emptyset$, $\mathcal{S} \leftarrow \emptyset$ \\
	$F \leftarrow Cost(\hat{D}^{(t)}, T^{(t-1)})$ \ \ // $F_{ij} = ||\hat{\vec{p}}_i^{(t)}-\hat{\vec{v}}, \vec{p}_j^{(t-1)}||_2$ \\
	\For{$i \leftarrow 1 \ to \ N$}{
	    $j \leftarrow \argmin_{j \notin \mathcal{S}} F_{ij}$ \\
	    // Class-wise distance threshold $\vec{\tau}_{c}$ \\
	    \If {$\vec{F}_{ij} \leq \vec{\tau}_{c}$}{
            // Associate with tracked object \\
            $a_{i}^{(t)} \leftarrow 0$ \\
            $T^{(t)} \leftarrow T^{(t)} \cup \{(\hat{D}_{i}^{(t)}, id_j^{(t-1)}, a_{i}^{(t)})\}$ \\
            $\mathcal{S} \leftarrow \mathcal{S} \cup \{j\}$ 
            // Mark track $j$ as matched\\
	    }
	    \Else {
	        // Initialize a new track \\
	        $a_{i}^{(t)} \leftarrow 0$ \\
	        $T^{(t)} \leftarrow T^{(t)} \cup \{(\hat{D}_{i}^{(t)}, newID, a_{i}^{(t)})\}$ 
	    }
	}
	\For{$j \leftarrow 1 \ to \ M $} {
	    \If {$j \notin \mathcal{S}$}{ // Unmatched tracks \\
	        \If {$T.a_{j}^{(t-1)} < A$}{
	            $T.a_{j}^{(t)} \leftarrow T.a_{j}^{(t-1)} + 1$ \\
	            $T.p_{j}^{(t)} \leftarrow T.p_{j}^{(t-1)} + T.v_{j}^{(t-1)}$ // Update the center location \\ 
	            $T^{(t)} \leftarrow T^{(t)} \cup \{T^{(t-1)}_j\}$  \\ 
	        }
	    }
	}
    \textbf{Return} $T^{(t)}$
\end{algorithm}

\section{Implementation Details}
Our implementation is based on the open-sourced code of CBGS~\cite{zhu2019classbalanced}\footnote{\url{https://github.com/poodarchu/Det3D}}.
CBGS provides implementations of PointPillars~\cite{pillar} and VoxelNet~\cite{voxelnet} on nuScenes. 
For Waymo experiments, we use the same architecture for VoxelNet and increases the output stride to 1 for PointPillars~\cite{pillar} following the dataset's reference implementation\footnote{\url{https://github.com/tensorflow/lingvo/tree/master/lingvo/tasks/car}}. 

A common practice~\cite{zhu2019classbalanced, yang20203dssd, vora2019pointpainting, caesar2019nuscenes} in nuScenes is to transform and merge the Lidar points of non-annotated frames into its following annotated frame.
This produces a denser point-cloud and enables a more reasonable velocity estimation.
We follow this practice in all nuScenes experiments. 

For data augmentation, we use random flipping along both $X$ and $Y$ axis,
and global scaling with a random factor from $[0.95, 1.05]$.
We use a random global rotation between $[-\pi/8, \pi/8]$ for nuScenes~\cite{zhu2019classbalanced} and $[-\pi/4, \pi/4]$ for Waymo~\cite{pvrcnn}.
We also use the ground-truth sampling~\cite{yan2018second} on nuScenes to deal with the long tail class distribution, which copies and pastes points inside an annotated box from one frame to another frame.  

For nuScenes dataset, we follow CBGS~\cite{zhu2019classbalanced} to optimize the model using AdamW~\cite{adamW} optimizer with one-cycle learning rate policy~\cite{one_cycle},
with max learning rate 1e-3, weight decay 0.01, and momentum $0.85$ to $0.95$.
We train the models with batch size 16 for 20 epochs on 4 V100 GPUs.

We use the same training schedule for Waymo models except a learning rate 3e-3, and we train the model for 30 epochs following PV-RCNN~\cite{pvrcnn}. 
To save computation on large scale Waymo dataset, we finetune the model for 6 epochs with second stage refinement modules for various ablation studies. 
All ablation experiments are conducted in this same setting.

For the nuScenes test set submission, we use a input grid size of $0.075m \times 0.075m$ and add two separate deformable convolution layers~\cite{dai2017deformable} in the detection head to learn different features for classification and regression.  
This improves CenterPoint-Voxel's performance from $64.8$ NDS to $65.4$ NDS on nuScenes validation. 
For the nuScenes tracking benchmark, we submit our best CenterPoint-Voxel model with flip testing, which yields a result of $66.5$ AMOTA on nuScenes validation.  

\section{nuScenes Performance across classes}
We show per-class comparisons with state-of-the-art methods in \reftab{nusc_per_cls}. 

\begin{table*}[t]
\centering 
\begin{tabular}{l@{\ \ \ }c@{\ \ \ }c@{\ \ \ }c@{\ \ \ }c@{\ \ \ }c@{\ \ \ }c@{\ \ \ }c@{\ \ \ }c@{\ \ \ }c@{\ \ \ }c@{\ \ \ }c@{\ \ \ }c}
  \toprule 
  Method & mAP & NDS & Car & Truck & Bus & Trailer & CV & Ped & Motor & Bicycle & TC & Barrier \\ 
 \cmidrule(r){1-1}
 \cmidrule(r){2-3}
 \cmidrule(){4-13}
  WYSIWYG \cite{hu2019exploiting} & 35.0 & 41.9 & 79.1 & 30.4 & 46.6 & 40.1 & 7.1 & 65.0 & 18.2 & 0.1 & 28.8 & 34.7\\ 
  PointPillars \cite{pillar} & 30.5 & 45.3 & 68.4 & 23.0 & 28.2 & 23.4 & 4.1 & 59.7 & 27.4 & 1.1 & 30.8 & 38.9\\ 
  PointPainting \cite{vora2019pointpainting} & 46.4 & 58.1 & 77.9 & 35.8 & 36.2 & 37.3 & 15.8 & 73.3 & 41.5 & 24.1 & 62.4 & 60.2\\ 
  CVCNet \cite{chen2020view} & 55.3 & 64.4 & 82.7 & 46.1 & 46.6 & 49.4 & \textbf{22.6} & 79.8 & \textbf{59.1} & \textbf{31.4} & 65.6 & 69.6 \\ 
  PMPNet \cite{yin2020Lidarbased} & 45.4 & 53.1 & 79.7 & 33.6 & 47.1 & 43.1 & 18.1 & 76.5 & 40.7 & 7.9 & 58.8 & 48.8\\ 
  SSN \cite{zhu2020ssn} & 46.4 & 58.1 & 80.7 & 37.5 & 39.9 & 43.9 & 14.6 & 72.3 & 43.7& 20.1 & 54.2 & 56.3 \\ 
  CBGS \cite{zhu2019classbalanced} & 52.8 & 63.3 & 81.1 & 48.5 & 54.9 & 42.9 & 10.5 & 80.1 & 51.5 & 22.3 & 70.9 & 65.7\\ 
  Ours & \textbf{58.0} & \textbf{65.5} & \textbf{84.6} & \textbf{51.0} & \textbf{60.2} & \textbf{53.2} & 17.5 & \textbf{83.4} & 53.7 & 28.7 & \textbf{76.7} & \textbf{70.9} \\  
  \bottomrule
 \end{tabular}
\caption{State-of-the-art comparisons for 3D detection on nuScenes test set. We show the NDS, mAP, and mAP for each class. Abbreviations: construction vehicle (CV), pedestrian (Ped), motorcycle (Motor), and traffic cone (TC).}
 \lbltab{nusc_per_cls}
\end{table*}

\begin{table}[t]
\vspace{-2mm}
\begin{center}
\begin{tabular}{lcc}
  \toprule
   Method & mAP & NDS \\
  \midrule
  Baseline & 57.1 & 65.4  \\ 
  + PointPainting~\cite{vora2019pointpainting} & 62.7 & 68.0 \\ 
  + Flip Test & 64.9 & 69.4 \\ 
  + Rotation & 66.2 & 70.3 \\ 
  + Ensemble & 67.7& 71.4 \\ 
  + Filter Empty & 68.2 & 71.7 \\
  \bottomrule

\end{tabular}
\end{center}
\vspace{-5mm}
\caption{Ablation studies for 3D detection on nuScenes validation. 
}
\lbltab{nusc_challenge}
\end{table}

\section{nuScenes Detection Challenge}
As a general framework, CenterPoint is complementary to contemporary methods and was used by three of the top 4 entries in the NeurIPS 2020 nuScenes detection challenge.  
In this section, we describe the details of our winning submission which significantly improved 2019 challenge winner CBGS~\cite{zhu2019classbalanced} by $14.3$ mAP and $8.1$ NDS. 
We report some improved results in \reftab{nusc_challenge}. 
We use PointPainting~\cite{vora2019pointpainting} to annotate each lidar point with image-based instance segmentation results generated by a Cascade RCNN model trained on nuImages\footnote{acquired from \url{https://github.com/open-mmlab/mmdetection3d/tree/master/configs/nuimages}}. 
This improves the NDS from $65.4$ to $68.0$. 
We then perform two test-time augmentations including double flip testing and point-cloud rotation around the yaw axis. 
Specifically, we use [0\textdegree, $\pm$ 6.25\textdegree, $\pm$ 12.5\textdegree, $\pm$ 25\textdegree] for yaw rotations.
Theses test time augmentations improve the NDS from $68.0$ to $70.3$.
In the end, we ensemble five models with input grid size between $[0.05m, 0.05m]$ to $[0.15m, 0.15m]$ and filter out predictions with zero number of points, which yields our best results on nuScenes validation, with $68.2$ mAP and $71.7$ NDS.

\end{document}